\documentclass[pdflatex,sn-mathphys-num]{sn-jnl}% Math and Physical Sciences Numbered Reference Style 
\usepackage{graphicx}%
\usepackage{multirow}%
\usepackage{amsmath,amssymb,amsfonts}%
\usepackage{amsthm}%
\usepackage{mathrsfs}%
\usepackage[title]{appendix}%
\usepackage{xcolor}%
\usepackage{textcomp}%
\usepackage{manyfoot}%
\usepackage{booktabs}%
\usepackage{algorithm}%
\usepackage{algorithmicx}%
\usepackage{algpseudocode}%
\usepackage{listings}%
\usepackage{todonotes}
\usepackage[detect-all]{siunitx}
\usepackage{upgreek}
\usepackage{tabulary}
\usepackage{mathtools}
\usepackage{hyperref}

\theoremstyle{thmstyleone}%
\theoremstyle{thmstyletwo}%

\theoremstyle{thmstylethree}%

\begin{document}

\title[Temporal-Conditioned Normalizing Flows for Multivariate Time Series Anomaly Detection]{Temporal-Conditioned Normalizing Flows for Multivariate Time Series Anomaly Detection}


\author*[1]{\fnm{David} \sur{Baumgartner}}\email{david.baumgartner@ntnu.no}

\author[1]{\fnm{Helge} \sur{Langseth}}%\email{helge.langseth@ntnu.no}

\author[]{\fnm{Kenth} \sur{Engø-Monsen}}%\email{kenthe@gmail.com}

\author[1]{\fnm{Heri} \sur{Ramampiaro}}%\email{heri@ntnu.no}

\affil[1]{
\orgdiv{Department of Computer Science}, 
\orgname{Norwegian University of Science and Technology}%
%\orgaddress{\street{Holtermannsveien 2}, \city{Trondheim}, \postcode{7030}, \state{Trøndelag}, \country{Norway}}
}

\abstract{

This paper introduces temporal-conditioned normalizing flows (tcNF), a novel framework that addresses anomaly detection in time series data with accurate modeling of temporal dependencies and uncertainty. 
By conditioning normalizing flows on previous observations, tcNF effectively captures complex temporal dynamics and generates accurate probability distributions of expected behavior. 
This autoregressive approach enables robust anomaly detection by identifying low-probability events within the learned distribution.
We evaluate tcNF on diverse datasets, demonstrating good accuracy and robustness compared to existing methods. 
A comprehensive analysis of strengths and limitations and open-source code is provided to facilitate reproducibility and future research.
}

\keywords{Machine Learning, Generative Model, Time Series, Anomaly Detection}



\maketitle

\newcommand{\todoi}[1]{\todo[inline]{#1}}
\newcommand{\xPDF}{p_{\boldsymbol{X}}}
\newcommand{\xBold}{\boldsymbol{x}}
\newcommand{\uPDF}{p_{\boldsymbol{U}}}
\newcommand{\uBold}{\boldsymbol{u}}
\newcommand{\fBold}{\boldsymbol{f}}
\newcommand{\FBold}{\boldsymbol{F}}
\newcommand{\gBold}{\boldsymbol{g}}
\newcommand{\GBold}{\boldsymbol{G}}
\newcommand{\aBold}{\boldsymbol{a}}
\newcommand{\bBold}{\boldsymbol{b}}
\newcommand{\cBold}{\boldsymbol{c}}
\newcommand{\dBold}{\boldsymbol{d}}
\newcommand{\hBold}{\boldsymbol{h}}
\newcommand{\sBold}{\boldsymbol{s}}
\newcommand{\tBold}{\boldsymbol{t}}
\newcommand{\wBold}{\boldsymbol{w}}

\newcommand{\xd}{\xBold^{1:d}}
\newcommand{\xD}{\xBold^{d+1:D}}
\newcommand{\ud}{\uBold^{1:d}}
\newcommand{\uD}{\uBold^{d+1:D}}

\newcommand{\xtd}{\xBold_t^{1:d}}
\newcommand{\xtD}{\xBold_t^{d+1:D}}
\newcommand{\utd}{\uBold_t^{1:d}}
\newcommand{\utD}{\uBold_t^{d+1:D}}

\newcommand{\distribution}{\blacktriangle}
\newcommand{\reconstruction}{\blacksquare}
\newcommand{\distance}{\blacklozenge}
\newcommand{\forecasting}{\bigstar}
\newcommand{\trees}{\blacktriangleleft}

\newcommand{\deepl}[1]{\textcolor{red}{#1}}
\newcommand{\classic}[1]{\textcolor{blue}{#1}}
\newcommand{\outlier}[1]{\textcolor{black}{#1}}

\newcommand\footnoteref[1]{\protected@xdef\@thefnmark{\ref{#1}}\@footnotemark}

\section{Introduction}
\label{sec:introduction}

The increasing reliance on complex interconnected systems, from financial markets to industrial control systems, has made the detection of anomalous behavior more critical than ever\,\cite{chandola_anomaly_2009,zamanzadeh_darban_deep_2024}. 
Undetected and unhandled anomalies can lead to significant financial losses, system failures, and, in extreme cases, safety hazards. 
Identifying unusual patterns in vast amounts of multivariate time series data generated by these systems presents a significant challenge, requiring robust methods that are able to capture complex temporal dependencies and account for inherent uncertainties.

A crucial aspect is the inherent interdependence between individual time series\,\cite{tsay_multivariate_2014}.
As an example, when modeling the production-demand balance within a power grid, it is essential to consider not just the grid's past performance but also connected grids and other relevant measurements, as these can contribute to power spikes or, in severe cases, blackouts.
To illustrate further, consider a production line within a network of processing steps using sensor data.
A disturbance in one processing step will have repercussions on the sensor readings of neighboring processing steps, an effect that a univariate model would likely fail to capture.
Therefore, to provide accurate anomaly detection, the models must account for these dependencies.

So, what is anomaly detection?
Among many definitions of anomaly detection available in the literature, we adopt the one provided in\,\citet{chandola_anomaly_2009}, referring to anomaly detection as \textit{``the problem of finding patterns in data that do not conform to expected or normal behavior''}.
This definition is operational in that it suggests first to establish a model or a representation originating from \textit{``expected or normal behavior''} and then to compare new data points with this gold standard.
More importantly, it accounts for the need for robust methods for complex time series data, as mentioned above.

To address the challenges of anomaly detection in complex time series, we employ normalizing flows \cite{papamakarios_normalizing_2021}, a powerful class of generative models capable of learning complex probability distributions. 
By training a normalizing flow on normal data, we can efficiently compute the log-likelihood of new data points, enabling us to identify anomalies as low-likelihood observations.
%By training a normalizing flow on normal data, we can efficiently compute the log-likelihood, $\log p(\xBold)$, of new data points $\xBold$, enabling us to identify anomalies as low-likelihood observations.
Crucially, we leverage the ability to condition normalizing flows to capture temporal dependencies within the time series.

In summary, our main contributions are as follows.
\begin{itemize}
    \item We propose temporal-conditioned normalizing flows (tcNF), a novel probabilistic anomaly detection framework that explicitly models temporal dependencies in time series data.
    Our framework offers enhanced accuracy and robustness compared to existing methods.
    \item We suggest a framework using unsupervised learning, but can still use labels in selecting good solution candidates if available.
    The different complexity levels of the methods further provide insights into the model complexity necessary for a given sequence.
    \item We perform a thorough comparison with state-of-the-art methods using two synthetic benchmark suites (mTADS\,\cite{baumgartner_mtads_2023}) and five real-world datasets.
    Our evaluation reveals that our framework achieves competitive performance in many scenarios, while also exposing specific challenging sequence patterns.
    \item We provide a repository with code, test configurations, and full result tables to reproduce, extend or integrate our framework.
\end{itemize}

The rest of this paper is organized as follows. Sec.\,\ref{sec:background} introduces normalizing flows and the foundation of conditioned coupling layers. Sec.\,\ref{sec:relatedwork} discusses recent developments related to conditioning or anomaly detection with normalizing flows. Sec.\,\ref{sec:method} covers temporal-conditioned coupling layers in the context of time series data. Sec.\,\ref{sec:experiment} describes the experiment, the datasets, and the evaluation, including the results for each dataset suite. Finally, Sec.\,\ref{sec:conclusion} concludes the paper and outlines possible future work.

\section{Background and Preliminaries}
\label{sec:background}

\subsection{Normalizing Flows}

In the context of generative modeling, normalizing flows offer a powerful technique for density estimation\,\cite{papamakarios_normalizing_2021}. 
The fundamental principle behind normalizing flows is to transform samples from a simple, well-defined distribution, such as a standard normal distribution, into a complex, unknown probability distribution. 
This transformation is learned through a series of invertible mappings, which allows for both evaluating the likelihood of observed data and generating new samples from the learned distribution.
\citet{papamakarios_normalizing_2021} define normalizing flows as a method that creates complex probability distributions $\xPDF$ by passing random variables $\uBold \in \mathbb{R}^{D}$, drawn from a simple base distribution $\uPDF$, commonly a Gaussian distribution, through $N$ nonlinear but invertible transformations (bijections).
Even with a simple base distribution, a sufficiently flexible transformation $\FBold$ can produce a complex distribution for the transformed variable $\xBold$.
To achieve such a transformation, normalizing flows commonly combine a sequence of simpler transformations: % together, defined as
\begin{equation}
    \xBold = \FBold(\uBold) = \fBold_N\left(\fBold_{N-1}\left(\cdots \fBold_2\left(\fBold_1(\uBold)\right)\right)\right), %\text{ where } \uBold \sim \uPDF.
    \label{eq:NF-chain}
\end{equation}
where $\uBold \sim \uPDF$.
The forward propagation draws $\uBold$ from $\uPDF$ and computes $\xBold = \FBold(\uBold)$. 
The inverse function $\uBold = \GBold(\xBold) = \FBold^{-1}(\xBold)$, maps $\xBold$ back to the base distribution, effectively \textit{normalizing} the data distribution.
When setting $N=1$ for simplicity, the change-of-variables formula gives
\begin{equation}
    \xPDF(\xBold) = \uPDF(\gBold(\xBold)) \vert \det \mathbf{J}(\gBold)(\xBold) \vert = \uPDF(\uBold) \vert \det \mathbf{J}(\fBold)(\uBold) \vert^{-1},
    \label{eq:NF-sample}
\end{equation}
where $\mathbf{J}(\gBold)(\xBold) = \frac{\partial\gBold}{\partial\xBold} \vert_{\xBold}$ is the Jacobian matrix of $\gBold = \fBold^{-1}$ evaluated at $\xBold$. 
This generalizes directly to $N>1$.
Taking the $\log$ on both sides and for multiple transformations results in 
\begin{equation}
    \log \xPDF(\xBold) = \log \uPDF(\GBold(\xBold)) + \sum_{i=1}^{N} \log \vert \det \mathbf{J}(\gBold_i)(\xBold_i)\vert,
    \label{eq:NF-optimisation-goal}
\end{equation}
where $\xBold_i = \gBold_i\left(\cdots \gBold_1\left(\xBold\right)\right)$ is the $i$-th intermediate output of the sequences of inverted transformations. 
The Jacobian determinant in Eq.\,\ref{eq:NF-optimisation-goal} of the transformation function $\gBold_i$ is a central point in each transformation.
Hence, a well-designed $\gBold_i$ ensures efficient computation and evaluation, commonly resulting in a lower triangular Jacobian, for which the determinant is comparatively efficient.
Building on the introduced concepts, we can efficiently evaluate the log density of $\log \xPDF(\xBold)$.

Thus, a normalizing flow is generative where we can both efficiently generate new data (Eq.\,\ref{eq:NF-chain}) and calculate the log-likelihood of any data point (Eq.\,\ref{eq:NF-optimisation-goal}). 
This contests other popular generative models like diffusion, which do not provide an exact calculation of likelihoods.

\subsection{Coupling Layer}
\label{subsec:coupling-layer}

\begin{figure}
    \centering
    \includegraphics[width=1\textwidth]{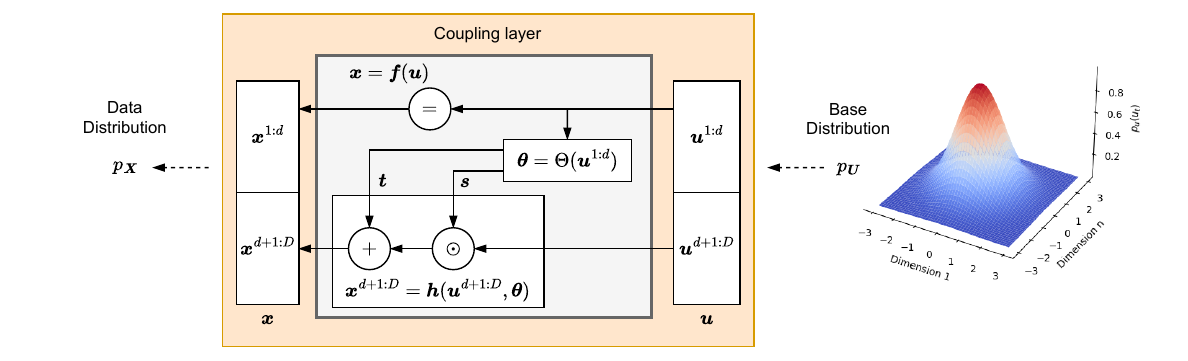}
    \caption{Forward propagation of one coupling layer from the base distribution to the data distribution.}
    \label{fig:base-coupling-layer}
\end{figure}

Coupling layers were outlined by \citet{dinh_density_2017} as the transformation in a normalizing flow, which models dependencies between dimensions using non-linear functions. 
In general a coupling layer partitions the input $\xBold$ into two subspaces, $(\xd, \xD) \in \mathbb{R}^d \times \mathbb{R}^{D-d}$, where $d \in \{1, 2, \cdots, D-1\}$.
The output $\uBold$ is equally partitioned as $(\ud, \uD) \in \mathbb{R}^d \times \mathbb{R}^{D-d}$.
Each coupling layer contains a transformation function $\hBold(\cdot, \boldsymbol{\theta}) : \mathbb{R}^d \rightarrow \mathbb{R}^{D-d}$, which is parameterized by a \textit{conditioner} function $\boldsymbol{\theta} = \Theta(\uBold^{1:d})$. 
Those functions define the basics of one coupling layer $\fBold_i : \mathbb{R}^D \rightarrow \mathbb{R}^D$ in a chain of transformations as
\begin{equation}
    \begin{split}
        \xd &= \ud \\
        \xD &= \hBold (\uD, \Theta(\ud)).
    \end{split}
\end{equation}
This couples $\ud$ into $\uD$ through the functions $\hBold(\cdot)$ and the conditioner $\Theta(\cdot)$. 
An \textit{affine} coupling layer is characterized as
\begin{equation}
    \xD = \hBold(\uD, \Theta (\ud)) = \uD \odot \exp(\sBold) + \tBold,    
    \label{eq:base-affine-coupling-layer}
\end{equation}
where $(\sBold, \tBold) = \boldsymbol{\theta} = \Theta (\ud) \in (\mathbb{R}^{D-d}, \mathbb{R}^{D-d})$ provide scale and translation respectively, and $\odot$ is the Hadamard product, see Fig.\,\ref{fig:base-coupling-layer}.
The conditioner function $\Theta(\cdot)$ is typically implemented as a deep neural network, providing the transformation parameters. 
This affine coupling layer $\fBold_i$ is invertible, and the Jacobian is guaranteed to be block triangular.
Using a chain of multiple coupling layers, the normalizing flow reaches full information exchange between the first $\uBold^{1:d}$ and the second $\uBold^{d+1:D}$ feature group by swapping the conditioning direction or the feature positions.
It is recommended to use a hyperbolic tangent on the conditioning output $\sBold$ with a learnable scalar vector to prevent the feature vector from collapsing\,\cite{dinh_density_2017,papamakarios_neural_2019}.

\subsection{Conditioned Coupling Layer}
\label{subsec:conditioned-coupling-layer}

%As pointed out by\,\cite{rasul_multivariate_2020,prenger_waveglow_2019}, the conditioner function $\Theta$ is not required to be invertible, enabling a flexible and general conditioning of the NFs. 
%We use $\boldsymbol{w} \in \mathbb{R}^L$, where $L$ is fixed but independent from the general dimension $D$ of the feature vector $\xBold$, as a condition which can be raw data or the output of another model, which can be jointly optimized with the NF. 
%Commonly, the conditioner function $\Theta$ utilizes this condition as additional input in providing $\boldsymbol{\theta}$ and is written as $\boldsymbol{\theta} = \Theta(\ud, \boldsymbol{w})$. 
%We refer to this as a conditioned coupling layer, allowing a conditional joint distribution $\xPDF(\xBold | \boldsymbol{w})$. 

We will use coupling layers to analyze time series data, and use $\Theta(\cdot)$ to condition the normalizing flow for $\xBold$, e.g., on previous observations $\xBold_1, \xBold_2, \dots, \xBold_{t-1}$. 
In particular, we define $\boldsymbol{w} \in \mathbb{R}^L$ as the summary of additional information, e.g., the concatenation of previous observations or some learned aggregation. 
Note that $L$, the dimensionality of $\boldsymbol{w}$, is a hyperparameter that we can optimize. 
Now, the conditioner function $\Theta(\cdot)$ utilizes this additional input as a condition and is written as $\boldsymbol{\theta} = \Theta(\ud, \boldsymbol{w})$. 
This approach is computationally feasible because, as pointed out by\,\cite{rasul_multivariate_2020,prenger_waveglow_2019}, the conditioner function $\Theta(\cdot)$ is not required to be invertible, enabling a flexible and general conditioning of the normalizing flows, see Fig.\,\ref{fig:timeflow-overview}.

\section{Related Work}
\label{sec:relatedwork}

In the following, we provide an overview of past work regarding normalizing flows, mainly in the context of time series applications.

\citet{guan_conditional_2023} propose an interesting approach using masked autoregressive flows with a negative log-likelihood optimization combined with a contrastive objective for a neighboring attention effect.
However, it is not clear to us whether the authors use input conditioning or transformation conditioning. 
Unfortunately, this aspect is not explicitly detailed in the publication and the implementation code is not publicly available. 
Also, direct results comparison is not possible as it seems this approach uses point adjustment strategies for the scoring, which is, as explained later, different from ours.

\citet{ning_anomaly_2023}, \citet{dai_graph_augmented_2021}, and \citet{zhou_label-free_2024} use a graph convolutional network and a graph attention network to model inter-dependencies, followed by a vanilla normalizing flow with extended input, resulting in input conditioning. 
\citet{brockmann_voraus-ad_2024}, on the other hand, changes the conditioner model from an MLP to a CNN model to enforce a sequential interconnection between the source channels and the target channels in each coupling step.
With \citet{rasul_multivariate_2020} and \citet{kang_traffic_2022}, we see the extension of normalizing flows towards fully conditional normalizing flows. 
That means the transformation functions, such as the layers in RealNVP\,\cite{dinh_density_2017}, are directly enriched with conditional information provided by an additional model outside the normalizing flow. 
This conditioning approach is versatile and not restricted to time series data only.
\citet{rasul_multivariate_2020} focuses on forecasting with LSTM and transformer models as encoders, in contrast to \citet{kang_traffic_2022}, which includes clustering to group the data points to perform anomaly detection in each group independently.
\citet{moon_multivariate_2023} uses the same conditioning solution but includes the current time step. 
This solution prevents the generation of new data points since the encoder needs the current data point to create the required conditioning for the flow.

\citet{zhao_ddanf_2024} combines a classic VAE with a normalizing flow (RealNVP), utilizing channel shuffling and embedding to reconstruct a sequence unsupervised. 
This approach, together with a loss based on three terms, integrates two related approaches, the log-likelihood and the reconstruction error, to do anomaly detection.
This work highlights a classic combination to get to a log-likelihood, but locks the normalizing flow and its generative capabilities away.
\citet{kang_transformer-based_2024} introduces two versions of variable temporal transformers, VTT-SAT and VTT-PAT, which highlights a new way of processing multivariate time series data.
It is an autoencoder approach where the anomaly detection originates from the reconstruction error and lacks the possibility of generating data points.

\section{Method}
\label{sec:method}

\begin{figure*}
    \centering
    \includegraphics[width=0.9\textwidth]{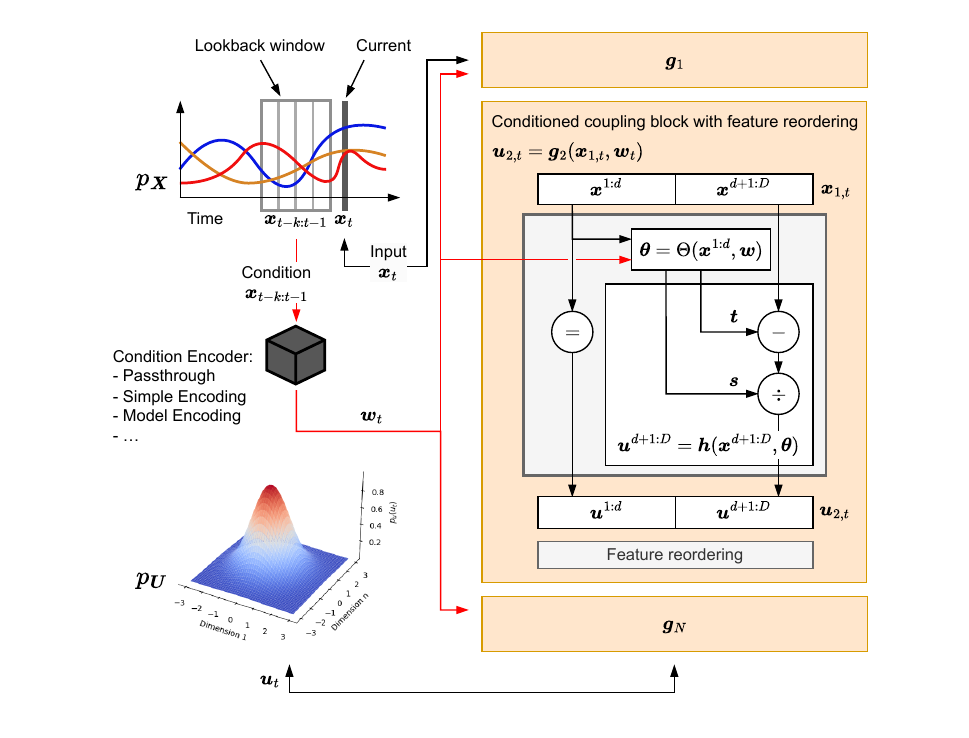}
    \caption{Overview of a temporal-conditioned coupling layer, drawn as normalization (reverse process) of the data distribution $\uBold_{2,t} = \gBold(\xBold_{1,t}, \wBold_t)$. We drop the coupling layer index $i = 1,\dots,N$ and the time index $t$ from $\xBold_{i,t}$ and $\uBold_{i,t}$ whenever it is clear from the context.
    The diagram illustrates how the conditioning information (red arrows) is added to the conditioner function expressed as $\wBold_t$. 
    After one such transformation, it is required to swap the two slices, such as $\xBold_{2,t}^{1:d} = \uBold_{2,t}^{d+1:D}$ and $\xBold_{2,t}^{d+1:D} = \uBold_{2,t}^{1:d}$, to achieve a full coupling of the input as the next $\xBold_t$ of the following layer.}
    \label{fig:timeflow-overview}
\end{figure*}

In Sec.\,\ref{sec:background}, we introduced normalizing flows as a technique for density estimation. 
In the following, we focus on anomaly detection by learning expected data behavior unsupervised.
We use this model further to calculate the likelihood of new data points.
The likelihood defines whether a point fits the normal behavior or not by applying a threshold based on a fixed value or extracted from some analysis after fitting the model, such as from the AUC-ROC metric.

\subsection{Temporal-Conditioned Coupling Layer}
\label{subsec:temporal-conditioned-coupling-layer}

The concept of conditioned coupling layers in Sec.\,\ref{subsec:conditioned-coupling-layer} provides a flexible framework with a strong impact on the transformation $\hBold(\cdot)$.
Similar to\,\cite{rasul_multivariate_2020}, we want to have temporal conditioning to respect the time and historical behavior in the transformation steps of the normalizing flow and call it a temporal-conditioned coupling layer. 
However, unlike\,\cite{rasul_multivariate_2020}, our aim is to do anomaly detection, and we are therefore investigating the use of temporal conditioners in that context.
Fig.\,\ref{fig:timeflow-overview} visualizes how we utilize temporal conditioning to enrich the conditioner function $\Theta(\cdot)$ with information from previous data points.
Notice how the conditioner function utilizes the immediate history represented by the $k$ last observations.
By employing temporal conditioning, we get an instant-based autoregressive effect over time. 
The key is to use a slice of previous timesteps, either in its raw form as a passthrough or enhanced via an encoder. 
Therefore, the conditioner in each coupling layer learns to utilize the available historical information.
Most commonly used temporal encoders are recurrent models\,\cite{rasul_multivariate_2020,moon_multivariate_2023,kang_traffic_2022}, which create embeddings as conditioning information that represent behavior over the previous observations.

We define a multivariate time series as a sequence $\left[\xBold_1, \cdots, \xBold_T\right]$, each $\xBold_t \in \mathbb{R}^D$, where $t$ is the time index $t \in [ 1,\cdots,T ]$. 
This allows us to learn a conditional probability distribution $\xPDF(\xBold_t | \xBold_{t-k:t-1})$ depending on the behavior of the current and previous timesteps.
The historical slice $\xBold_{t-k:t-1}$ refers to the matching input vector $\xBold_t$ at $t$ and is defined as $\wBold_t \in \mathbb{R}^L$. 
That way, $\Theta(\cdot)$ can efficiently utilize the conditioning information and defines one temporal-conditioned coupling layer as
\begin{equation}
    \label{eq:temporal-conditioned-coupling-layer}
    \begin{split}
        \xtd &= \utd \\
        \xtD &= \boldsymbol{h} (\utD, \Theta (\utd, \wBold_t)).
    \end{split}
\end{equation}
To regularize the model, we used the same $\wBold_t$ for all $\gBold_i, i=1,\dots,N$.
We write then the optimization objective and negative log-likelihood loss function based on Eq.\,\ref{eq:NF-optimisation-goal} for this temporal-conditioned normalizing flow as
\begin{equation}
    \label{eq:tcNF-optimization-objective}
    \mathcal{L}_t(\xBold_t, \wBold_t) = - \left[ \log \uPDF(\GBold(\xBold_t | \wBold_t)) 
    + 
    \sum_{i=1}^{N} \log \vert \det \mathbf{J}(\gBold_i)(\xBold_i) \vert \right],
\end{equation}
where $\log \vert \det \mathbf{J}(\gBold_i)(\xBold_i)\vert = \log \vert \exp (\mathrm{sum} (\sBold_i)) \vert$ and the average loss is $\mathcal{L} = \frac{1}{T} \sum_{t=1}^{T} \mathcal{L}_t$.
We call this framework temporal-conditioned normalizing flows (tcNF).

Our experiments focus on examining efficient encodings of previous observations, with options ranging from simple passthrough to deep-learning encoder models. 
The following describes them briefly. 
Note that this method can utilize any network or encoding as conditioning.

\subsubsection{Basic Temporal-Conditioned Coupling Layer (tcNF-base)}
\label{subsubsec:basic-temporal-conditioned-coupling-layer}

The basic temporal-conditioned coupling layer operates on a window of timesteps. 
Here, $\wBold_t$ represents the input to the conditioning mechanism at time $t$, defined as a sequence of measurements $\xBold_{t-k:t-1}$, where $k \geq 1$ is a hyperparameter controlling the length of the lookback window.
The conditioner function in each coupling layer is responsible for directly learning important features from the provided information.

\subsubsection{Encoded Temporal-Conditioned Coupling Layer}
\label{subsubesec:encoded-temporal-conditioned-coupling-layer}

The encoded temporal-conditioned coupling layer covers the complete family of methods that operate on a fixed or learnable encoding of previous timesteps.

A \textbf{Basic Encoded Temporal-Conditioned Coupling Layer} utilizes an algorithm-based encoding of the previous observations. 
We define this as $\wBold_t = \texttt{encode}(\xBold_{t-k:t-1})$ where $\texttt{encode}(\cdot)$ is a function that encodes the given information and may compress it to a different dimensionality as the conditioning.
In this case, the conditioner function will learn the important features after a fixed feature extraction.

A \textbf{Batched Temporal-Conditioned Coupling Layer (tcNF-mlp, tcNF-cnn, tcNF-stateless)} uses a learnable feed-forward, convolution, or LSTM model instead of a fixed algorithm to encode a given slice of previous observations, which is learned end-to-end with the normalizing flow model. 
That trainable model encodes a given slice over the previous observations independently of other overlapping slices. 
Each slice is paired to $\xBold_t$ as $\wBold_t = \texttt{encoder}(\xBold_{t-k:t-1})$, where $\texttt{encoder}(\cdot)$ is a tunable encoder model that connects multiple timesteps in the slice's scope, expressing connections and correlations between timesteps and channels.
This is similar to the general concept of the transformer-conditioned RealNVP in \cite{rasul_multivariate_2020}.

A \textbf{Stateful LSTM Temporal-Conditioned Coupling Layer (tcNF-stateful)} follows the design of the RNN-conditioned RealNVP in\,\cite{rasul_multivariate_2020}, where the LSTM network is stateful, meaning that the state is handed over from timestep to timestep.
It is expressed, therefore, similar to the batched version but without batching capabilities on the previous observation slice as $\wBold_t = \texttt{encoder}_{\mathrm{ful}}(\xBold_{t-1})$ because there exists a sequential dependency between each step. 
Training this model is more time-consuming because of the sequential processing of each timestep and the lack of batch processing, which increases the training time.

\section{Experiments}
\label{sec:experiment}

In this section, we conduct a comprehensive set of experiments to validate the effectiveness of the proposed tcNF framework. The code for reproducing the results is available online along with the experiment results\,\footnote{\url{https://github.com/2er0/HistCondFlow}}. 
The reported experiments were conducted on a computation cluster with different GPUs ranging from P100 to H100.

To demonstrate the impact of temporal-conditioning of normalizing flows, we utilize the methods presented in Sec.\,\ref{subsec:temporal-conditioned-coupling-layer} with increasing complexity to demonstrate that simpler models are to be preferred on not overly complex datasets. 
Further, we use the covariance matrix adaptation evolution strategy (CMA-ES)\,\cite{auger_restart_2005} for the hyperparameter optimization. 
Sec.\,\ref{subsec:hyperparameter-and-evaluation} explains the different optimization goals based on data availability to achieve a directed parameter search. 

\subsection{Datasets}
\label{subsec:dataset}

To test the capabilities of the proposed framework, we use a diverse set of sequences with various levels of controlled environments to show the performance and capabilities of tcNF. 
\citet{baumgartner_mtads_2023} introduced two benchmark suites, collectively referred to as mTADS. The first suite simulates a highly controlled environment with 70 synthetic sequences representing diverse base sequence types and anomalies (FSB).
The second suite is semi-realistic, containing four extensively long sequences with highly interconnected channels and dependencies introduced during the generation process (SRB).
Together, these benchmark suites provide the possibility to assess the general performance under both controlled and semi-realistic conditions.

Evaluating the real-world capabilities of tcNF and show how well the framework of methods generalizes requires real-world data. 
We selected the following datasets:
\begin{itemize}
\item \textit{Secure Water Treatment} (SWaT): a dataset containing sensor data from a physical system, where we use the \textit{v1} training sequence and the \textit{v0} test sequence; 
\item \textit{CalIt2}: a dataset that was created by counting the number of people entering and leaving a building;
\item \textit{GHL}: is a dataset from the cyber security domain containing simulated attacks on a simulated industrial system;
\item \textit{Metro}: a dataset containing hourly traffic volume information and weather features;
\item \textit{Server Machine Dataset} (SMD): a dataset from KDD 2019 containing server measurements.
\end{itemize}
Most original publications, except Metro, provide test sequences with flagged anomalies. 
In this case, we rely on the anomaly flagging provided in this repository\,\footnote{\url{https://timeeval.github.io/evaluation-paper/notebooks/Datasets.html}} with the publication of\,\citet{schmidl_anomaly_2022}.
Tab.\,\ref{tab:dataset-stats} summarizes the characteristics of the datasets. 

\begin{table}[]
    \centering
    \caption{Description of the datasets. (\%) is the percentage of anomalous data points in the dataset.}
    \begin{tabular*}{\textwidth}{@{\extracolsep{\fill}}lrrrrr}
    Dataset & \# Datasets & Variables & Train & Test & Anomalies (\%) \\
    \midrule
    FSB\,\cite{baumgartner_mtads_2023} & 70 & \numrange{2}{10} & \multicolumn{2}{c}{\numrange{100}{10000}} & \numrange{0.09}{15.00} \\
    SRB\,\cite{baumgartner_mtads_2023} & 4 & 4 & \multicolumn{2}{c}{\numrange{199000}{999000}} & \numrange{0.01}{0.03} \\
    \midrule
    SWaT\,\cite{goh_dataset_2017} & 1 & 51 & 449919 & 449919 & $12.15$ \\
    CalIt2\,\cite{hutchins_calit2_2006} & 1 & 2 & $4032$ & $1008$ & $3.17$ \\
    GHL\,\cite{filonov_multivariate_2016} & 48 & 16 & $1535118$ & $200001$ & \numrange{0.39}{0.42} \\
    Metro\,\cite{hogue_metro_2019} & 1 & 5 & $38563$ & $9641$ & $0.13$ \\
    SMD\,\cite{su_robust_2019} & 28 & 38 & \numrange{23687}{28743} & \numrange{23687}{28743} & \numrange{0.42}{15.65} \\
    \end{tabular*}
    \label{tab:dataset-stats}
\end{table}

\subsection{Dataset Preparation}
\label{subsec:dataset-preparation}

Normalizing flows require normalized and non-categorical inputs, where we use min-max normalization, resulting in the range between \numrange{-1}{1}. 
The initially proposed coupling layers\,\cite{dinh_density_2017} possess requirements by design which are further relevant for our implementation. 
First, we set $d$ to $D/2$, which requires $D$ to be an even number. 
Each transformation modifies one-half of the features per step. 
Hence, an even number of features is needed to reach full information exchange and to keep the implementation clean.
Therefore, we add one channel with a constant $0.5$ in the case of an odd number of channels to have an even number of channels.
Channels containing only zeros get replaced with the value $0.5$ to prevent the features from collapsing. 

We split the training data into training and validation at a ratio of $80:20$, with different selection types based on the method.
If the method allows batching, then five random sections of the training data are selected and expanded to cover 20\% of the training data for validation. 
In the case of sequential training dependency, we use the last 20\% of the training data as validation. 
In both cases, we exclude additional data from the 80\% training portion based on the lookback window to prevent information from leaking into the validation or training set. 
Independent test sequences exist for each train sequence and need no further handling. 

\subsection{Hyperparameter optimization and evaluation}
\label{subsec:hyperparameter-and-evaluation}

The evaluation and hyperparameter optimization are interlinked and guided.
The two benchmark suites in mTADS (FSB \& SRB) have three sequences per dataset, each of which contains two sequences for training: one without anomalies for unsupervised training, and another with anomalies for supervised training.
The test sequence per dataset in the benchmark suites is solely used for the final evaluation of the performance and performance reporting of a chosen model from the optimization search.

To evaluate the detection performance, we utilize the scoring implementation in \cite{paparrizos_volume_2022}, which provides a broad range of standard metrics ranging from AUC-ROC to F1-Score and precision/recall.
We mainly analyze the performance on the following two non-binary parametric free metrics\,\cite{sorbo_navigating_2024}\,\footnote{Additional metric results are available with the code in the repository, such as AUC-PR or F1.}.
The first metric is AUC-ROC (AUC), which is still the standard for point anomaly detection. 
Second, VUS-ROC (VUS)\,\cite{paparrizos_volume_2022}, is a newly introduced metric that covers ranged anomalies via fuzziness around an anomaly. 
VUS requires one parameter that defines the max window size, which can be inferred or set. 
We don't use any adjustment strategy and report raw metric results for full transparency, which can lead to non-detections or missing detections in the beginning and after a flagged range.

The guided hyperparameter optimization relies on the Covariance Matrix Adaptation Evolution Strategy (CMA-ES)\,\cite{auger_restart_2005} via generations of populations that represent parameter combinations to find a good set of parameters. This search is guided via one of two loss functions defining a solution candidate's performance. In the special case of the mTADS datasets, we use a weighted combination of the AUC and VUS score with a balance of 30:70 on the evaluation dataset, meaning we give more weight to the VUS score and range anomalies. 
When we do not have a labeled evaluation dataset available, we use the validation loss to determine the fitting of the data. 
We tested various options to guide the hyperparameter search, and those two presented options produced the most satisfactory candidates. The first option focuses on performance and allows minimal larger models, whereas the second option leads to slimmer models and, therefore, less overfitting. 
Tab.\,\ref{tab:hyperparameters-cma-es-space} summarizes the tunable hyperparameters for each method and their respective search ranges.

\begin{table*}[]
    \centering
    \caption{The listed hyperparameters are optimized per dataset and explored with CMA-ES to find a good parameter set. $\Theta(\cdot)$ parameters are for the conditioner function in each coupling layer. Past and $\texttt{encoder}(\cdot)$ parameters are primarily for the encoder outside the normalizing flow. "-" indicates that this parameter is not part of a given method.}
    \label{tab:hyperparameters-cma-es-space}
    \small
    \begin{tabular*}{\textwidth}{@{\extracolsep{\fill}}lcccccc}
         Parameter & 
         {\hspace{-5mm}\rotatebox{40}{RealNVP}} &
         {\hspace{-5mm}\rotatebox{40}{tcNF-base}} & 
         {\hspace{-5mm}\rotatebox{40}{tcNF-mlp}} & 
         {\hspace{-5mm}\rotatebox{40}{tcNF-cnn}} & 
         {\hspace{-5mm}\rotatebox{40}{tcNF-stateless}} & 
         {\hspace{-5mm}\rotatebox{40}{tcNF-stateful}} \\
         \midrule
         Coupling layers & \multicolumn{6}{c}{For all models: \numrange{3}{20}} \\
         $\Theta(\cdot)$ multiplier & \multicolumn{6}{c}{For all models: \numrange{1}{50}} \\ 
         $\Theta(\cdot)$ layers & \multicolumn{6}{c}{For all models: \numrange{3}{8}} \\
         $\Theta(\cdot)$ dropout & \multicolumn{6}{c}{For all models: \numrange{0.1}{0.9}} \\
         $\Theta(\cdot)$ funnel factor & \multicolumn{6}{c}{For all models: \numrange{1}{10}} \\
         \midrule
         Look-back window & - & \multicolumn{5}{c}{FSB: \numrange{1}{50}, others: \numrange{1}{100}} \\
         $\texttt{encoder}(\cdot)$ layers & - & - & \numrange{3}{20} & \numrange{1}{5} & \numrange{1}{10} & \numrange{1}{10} \\ 
         $\texttt{encoder}(\cdot)$ dropout & - & - & \numrange{0.1}{0.9} & \numrange{0.1}{0.9} & \numrange{0.1}{0.9} & \numrange{0.1}{0.9} \\
         $\texttt{encoder}(\cdot)$ compression & - & - & \numrange{1}{20} & - & - & - \\
         $\texttt{encoder}(\cdot)$ size & - & - & - & \numrange{3}{7} & - & - \\
         $\texttt{encoder}(\cdot)$ max channel & - & - & - & \numrange{1}{20} & - & - \\
    \end{tabular*}
\end{table*}

All reported results are based on the chosen model from the hyperparameter search, which covers the variance of possible performance. 
Reported means and standard deviations are based on multiple datasets covered in the given group of datasets.
In case of the mTADS sequences, we execute every option as discussed above, aggregate the results, and report every result option in the repository. 

%%%%%%%%%%%%%%%%%%%%%%%%%%%%%%%%%%%%%%%%%%%%%%%%%%%%%%%%%%%%%%%%%%%%%%%%%%%%%%%%%%%%%%%%%%%%%%%%%%%%%%%%%
%%%%%%%%%%%%%%%%%%%%%%%%%%%%%%%%%%%%%%%%%%%%%%%%%%%%%%%%%%%%%%%%%%%%%%%%%%%%%%%%%%%%%%%%%%%%%%%%%%%%%%%%%
%%%%%%%%%%%%%%%%%%%%%%%%%%%%%%%%%%%%%%%%%%%%%%%%%%%%%%%%%%%%%%%%%%%%%%%%%%%%%%%%%%%%%%%%%%%%%%%%%%%%%%%%%
%%%%%%%%%%%%%%%%%%%%%%%%%%%%%%%%%%%%%%%%%%%%%%%%%%%%%%%%%%%%%%%%%%%%%%%%%%%%%%%%%%%%%%%%%%%%%%%%%%%%%%%%%
%%%%%%%%%%%%%%%%%%%%%%%%%%%%%%%%%%%%%%%%%%%%%%%%%%%%%%%%%%%%%%%%%%%%%%%%%%%%%%%%%%%%%%%%%%%%%%%%%%%%%%%%%
%%%%%%%%%%%%%%%%%%%%%%%%%%%%%%%%%%%%%%%%%%%%%%%%%%%%%%%%%%%%%%%%%%%%%%%%%%%%%%%%%%%%%%%%%%%%%%%%%%%%%%%%%
%%%%%%%%%%%%%%%%%%%%%%%%%%%%%%%%%%%%%%%%%%%%%%%%%%%%%%%%%%%%%%%%%%%%%%%%%%%%%%%%%%%%%%%%%%%%%%%%%%%%%%%%%
%%%%%%%%%%%%%%%%%%%%%%%%%%%%%%%%%%%%%%%%%%%%%%%%%%%%%%%%%%%%%%%%%%%%%%%%%%%%%%%%%%%%%%%%%%%%%%%%%%%%%%%%%

\subsection{Results: Fully Synthetic Benchmark Suite (FSB)}
\label{subsec:results-fsb}

\begin{figure}
    \centering
    \includegraphics[width=0.95\textwidth]{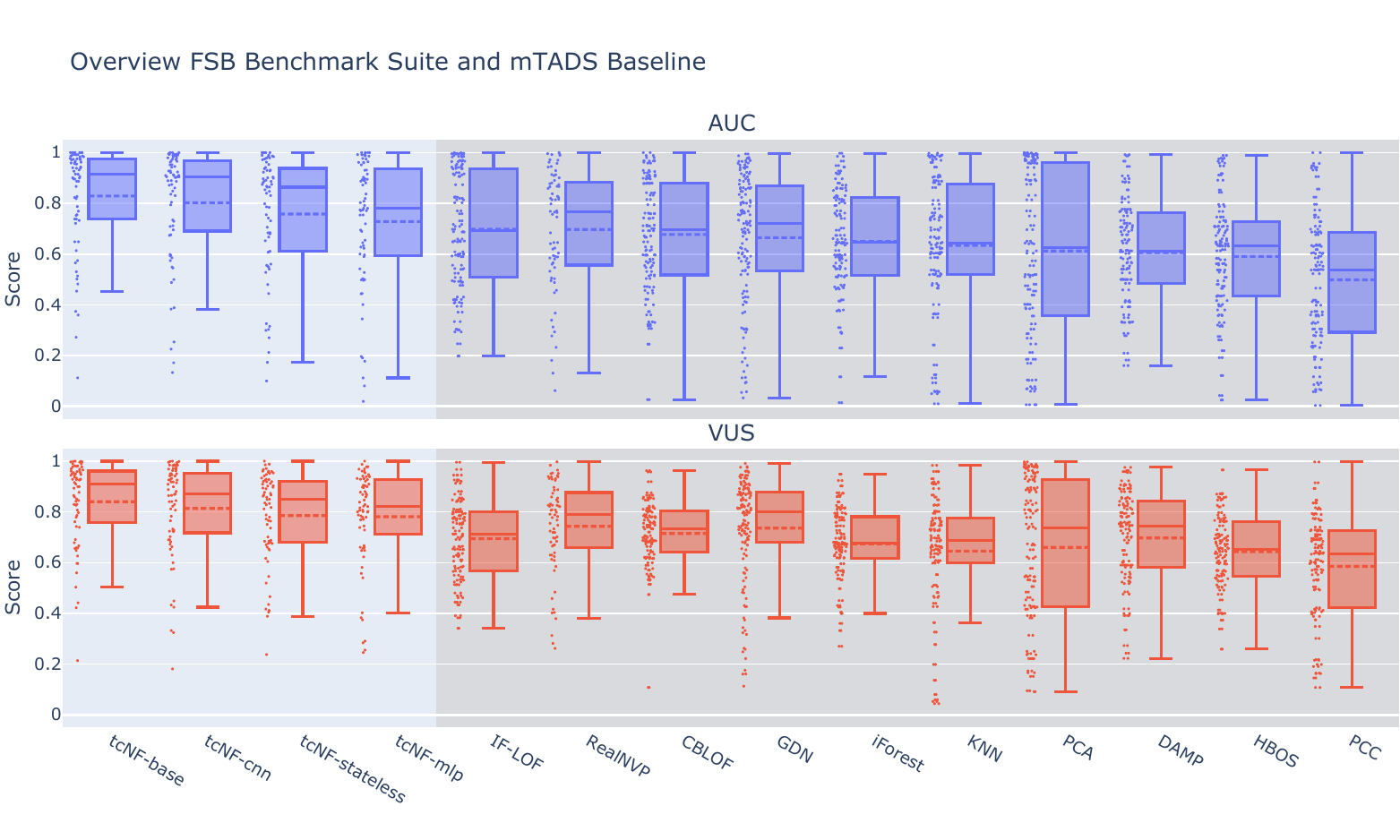}
    \caption{Comparison of anomaly detection performance on the FSB benchmark using AUC (top) and VUS (bottom) sorted by average performance on the AUC metric. The results demonstrate the competitive performance of the proposed tcNF models compared to established baselines, particularly in terms of RealNVP\,\cite{dinh_density_2017} as the unchanged baseline method.}
    \label{fig:fsb-boxplot-overview}
\end{figure}

The analyses of the results highlighted that our proposed method can outperform the base method and other existing methods, as can be seen in Fig.\,\ref{fig:fsb-boxplot-overview}. 
Results for other methods than tcNF and RealNVP are taken from\,\cite{baumgartner_mtads_2023}.
It is noticeable that our proposed base method, tcNF-base, shows solid performance on this benchmark suite, which is further understandable from the level of complexity in this suite.
Tab.\,\ref{tab:fsb-results} shows in further detail that sequences with smooth base behavior are preferred by our method. 
Analysing the results from the anomaly type indicates that different versions of our method perform better on different types\,\footnote{\label{note:see-repo}See repository for additional result tables and figures.}. 
For example, tcNF-base performs well on anomalies where the signal gets cut, and tcNF-cnn is favorable on platform, pattern, and amplitude changes. 
In general, anomalies such as mean change, variance, and trend are more difficult for our methods to detect.

\begin{table}
    \centering
    \caption{Comparing anomaly detection performance on the FSB benchmark from the sequence type using AUC (a) and VUS (b). The results show that our proposed tcNF models can handle most sequences well and shows difficulties on rapidly changing signals, demonstrating overall a competitive performance. Method categories: \deepl{red}: deep learning, \classic{blue}: classic machine learning, \outlier{black}: outlier detection, $\distribution$: distribution, $\reconstruction$: reconstruction, $\distance$: distance, $\forecasting$: forecasting and $\trees$: trees.}
    \label{tab:fsb-results}
    \resizebox{\textwidth}{!}{\begin{minipage}{\textwidth}
        \begin{tabular}{lcccccc}
    \multicolumn{7}{c}{(a) \textbf{AUC}} \\[0.2cm]
    & Sine (16) & ECG-CBF-RW (20) & Increasing (13) & Saw (2) & Wave (14) & Corr (5) \\
    Method &  &  &  &  &  &  \\
    \midrule
    \deepl{$\distribution$} tcNF-base & \bfseries 0.85 ± 0.15 & 0.69 ± 0.27 & \bfseries 0.97 ± 0.04 & 0.97 ± 0.01 & \bfseries 0.90 ± 0.10 & 0.70 ± 0.18 \\
    \deepl{$\distribution$} tcNF-mlp & 0.83 ± 0.14 & 0.59 ± 0.33 & 0.83 ± 0.12 & 0.98 ± 0.02 & 0.62 ± 0.26 & \bfseries 0.72 ± 0.17 \\
    \deepl{$\distribution$} tcNF-cnn & 0.83 ± 0.16 & 0.67 ± 0.31 & 0.89 ± 0.12 & 0.98 ± 0.01 & 0.87 ± 0.15 & 0.68 ± 0.19 \\
    \deepl{$\distribution$} tcNF-stateless & 0.78 ± 0.23 & 0.63 ± 0.30 & 0.88 ± 0.11 & 0.98 ± 0.01 & 0.76 ± 0.21 & 0.70 ± 0.15 \\
    \midrule
    \deepl{$\distribution$} RealNVP\,\cite{dinh_density_2017} & 0.79 ± 0.20 & 0.64 ± 0.26 & 0.80 ± 0.11 & 0.99 ± 0.00 & 0.54 ± 0.29 & 0.67 ± 0.16 \\
    \classic{$\reconstruction$} PCC\,\cite{shyu_novel_2003} & 0.37 ± 0.25 & 0.49 ± 0.25 & 0.50 ± 0.23 & 0.59 ± 0.06 & 0.45 ± 0.25 & 0.51 ± 0.09 \\
    \classic{$\distance$} HBOS\,\cite{goldstein_histogram-based_2012} & 0.55 ± 0.18 & 0.61 ± 0.21 & 0.32 ± 0.24 & 0.62 ± 0.04 & 0.60 ± 0.22 & 0.59 ± 0.06 \\
    \classic{$\distance$} KNN\,\cite{corchado_fourier_2009} & 0.77 ± 0.19 & 0.65 ± 0.26 & 0.09 ± 0.05 & \bfseries 0.99 ± 0.01 & 0.70 ± 0.16 & 0.51 ± 0.10 \\
    \classic{$\distance$} DAMP\,\cite{lu_damp_2023} & 0.61 ± 0.18 & 0.63 ± 0.16 & 0.73 ± 0.27 & 0.94 ± 0.06 & 0.45 ± 0.15 & 0.54 ± 0.09 \\
    \classic{$\reconstruction$} PCA & 0.81 ± 0.27 & 0.60 ± 0.33 & 0.83 ± 0.17 & \bfseries 0.99 ± 0.01 & 0.36 ± 0.20 & 0.51 ± 0.30 \\
    \deepl{$\forecasting$} Torsk\,\cite{heim_adaptive_2019} & 0.72 ± 0.24 & 0.75 ± 0.22 & 0.50 ± 0.00 & - & 0.79 ± 0.14 & - \\
    \deepl{$\forecasting$} GDN\,\cite{deng_graph_2021} & 0.82 ± 0.13 & 0.53 ± 0.31 & 0.75 ± 0.29 & 0.83 ± 0.10 & 0.70 ± 0.16 & 0.64 ± 0.10 \\
    \outlier{$\distance$} CBLOF\,\cite{he_discovering_2003} & 0.77 ± 0.15 & 0.69 ± 0.25 & 0.41 ± 0.25 & 0.84 ± 0.09 & 0.68 ± 0.16 & 0.57 ± 0.13 \\
    \outlier{$\distance$} COF\,\cite{tang_enhancing_2002} & 0.73 ± 0.14 & \bfseries 0.82 ± 0.17 & 0.17 ± 0.17 & 0.85 ± 0.00 & 0.70 ± 0.23 & 0.56 ± 0.10 \\
    \outlier{$\trees$} iForest\,\cite{liu_isolation_2008} & 0.71 ± 0.19 & 0.60 ± 0.25 & 0.56 ± 0.25 & 0.94 ± 0.01 & 0.61 ± 0.16 & 0.57 ± 0.04 \\
    \outlier{$\trees$} IF-LOF\,\cite{cheng_outlier_2019} & 0.71 ± 0.19 & 0.68 ± 0.26 & 0.64 ± 0.24 & 0.98 ± 0.02 & 0.71 ± 0.21 & 0.51 ± 0.10 \\
\end{tabular}

        \begin{tabular}{lcccccc}
    \multicolumn{7}{c}{(b) \textbf{VUS}} \\[0.2cm]
    & Sine (16) & ECG-CBF-RW (20) & Increasing (13) & Saw (2) & Wave (14) & Corr (5) \\
    Method &  &  &  &  &  &  \\
    \midrule
    \deepl{$\distribution$} tcNF-base & \bfseries 0.88 ± 0.10 & 0.72 ± 0.21 & \bfseries 0.97 ± 0.03 & 0.95 ± 0.03 & \bfseries 0.94 ± 0.06 & 0.85 ± 0.13 \\
    \deepl{$\distribution$} tcNF-mlp & 0.85 ± 0.11 & 0.66 ± 0.23 & 0.90 ± 0.08 & 0.95 ± 0.05 & 0.81 ± 0.20 & 0.84 ± 0.12 \\
    \deepl{$\distribution$} tcNF-cnn & 0.85 ± 0.14 & 0.67 ± 0.23 & 0.92 ± 0.09 & 0.96 ± 0.04 & 0.92 ± 0.11 & 0.85 ± 0.12 \\
    \deepl{$\distribution$} tcNF-stateless & 0.82 ± 0.13 & 0.65 ± 0.21 & 0.91 ± 0.07 & 0.96 ± 0.04 & 0.87 ± 0.10 & \bfseries 0.86 ± 0.12 \\
    \midrule
    \deepl{$\distribution$} RealNVP\,\cite{dinh_density_2017} & 0.80 ± 0.11 & 0.64 ± 0.20 & 0.88 ± 0.07 & 0.97 ± 0.02 & 0.75 ± 0.22 & 0.76 ± 0.10 \\
    \classic{$\reconstruction$} PCC\,\cite{shyu_novel_2003} & 0.55 ± 0.20 & 0.53 ± 0.22 & 0.70 ± 0.15 & 0.62 ± 0.01 & 0.54 ± 0.26 & 0.66 ± 0.09 \\
    \classic{$\distance$} HBOS\,\cite{goldstein_histogram-based_2012} & 0.64 ± 0.13 & 0.64 ± 0.15 & 0.47 ± 0.13 & 0.63 ± 0.02 & 0.71 ± 0.14 & 0.69 ± 0.09 \\
    \classic{$\distance$} KNN\,\cite{corchado_fourier_2009} & 0.78 ± 0.09 & 0.62 ± 0.17 & 0.14 ± 0.13 & 0.96 ± 0.00 & 0.76 ± 0.12 & 0.63 ± 0.13 \\
    \classic{$\distance$} DAMP\,\cite{lu_damp_2023} & 0.69 ± 0.16 & 0.70 ± 0.16 & 0.87 ± 0.15 & 0.93 ± 0.04 & 0.57 ± 0.19 & 0.64 ± 0.11 \\
    \classic{$\reconstruction$} PCA & 0.83 ± 0.19 & 0.61 ± 0.30 & 0.89 ± 0.09 & \bfseries 0.98 ± 0.00 & 0.42 ± 0.20 & 0.70 ± 0.17 \\
    \deepl{$\forecasting$} Torsk\,\cite{heim_adaptive_2019} & 0.75 ± 0.25 & 0.74 ± 0.26 & 0.51 ± 0.00 & - & 0.85 ± 0.16 & - \\
    \deepl{$\forecasting$} GDN\,\cite{deng_graph_2021} & 0.83 ± 0.10 & 0.56 ± 0.24 & 0.81 ± 0.22 & 0.82 ± 0.10 & 0.85 ± 0.09 & 0.76 ± 0.05 \\
    \outlier{$\distance$} CBLOF\,\cite{he_discovering_2003} & 0.77 ± 0.08 & 0.69 ± 0.11 & 0.52 ± 0.18 & 0.85 ± 0.08 & 0.78 ± 0.10 & 0.71 ± 0.09 \\
    \outlier{$\distance$} COF\,\cite{tang_enhancing_2002} & 0.74 ± 0.08 & \bfseries 0.77 ± 0.11 & 0.60 ± 0.24 & 0.86 ± 0.00 & 0.81 ± 0.07 & 0.69 ± 0.08 \\
    \outlier{$\trees$} iForest\,\cite{liu_isolation_2008} & 0.75 ± 0.10 & 0.56 ± 0.17 & 0.75 ± 0.05 & 0.91 ± 0.01 & 0.70 ± 0.12 & 0.67 ± 0.08 \\
    \outlier{$\trees$} IF-LOF\,\cite{cheng_outlier_2019} & 0.74 ± 0.10 & 0.61 ± 0.18 & 0.75 ± 0.08 & 0.94 ± 0.03 & 0.79 ± 0.10 & 0.60 ± 0.14 \\
\end{tabular}

    \end{minipage}}
    \vspace{-2em}
\end{table}

\begin{figure}
    \vspace{-1em}
    \includegraphics[width=\textwidth]{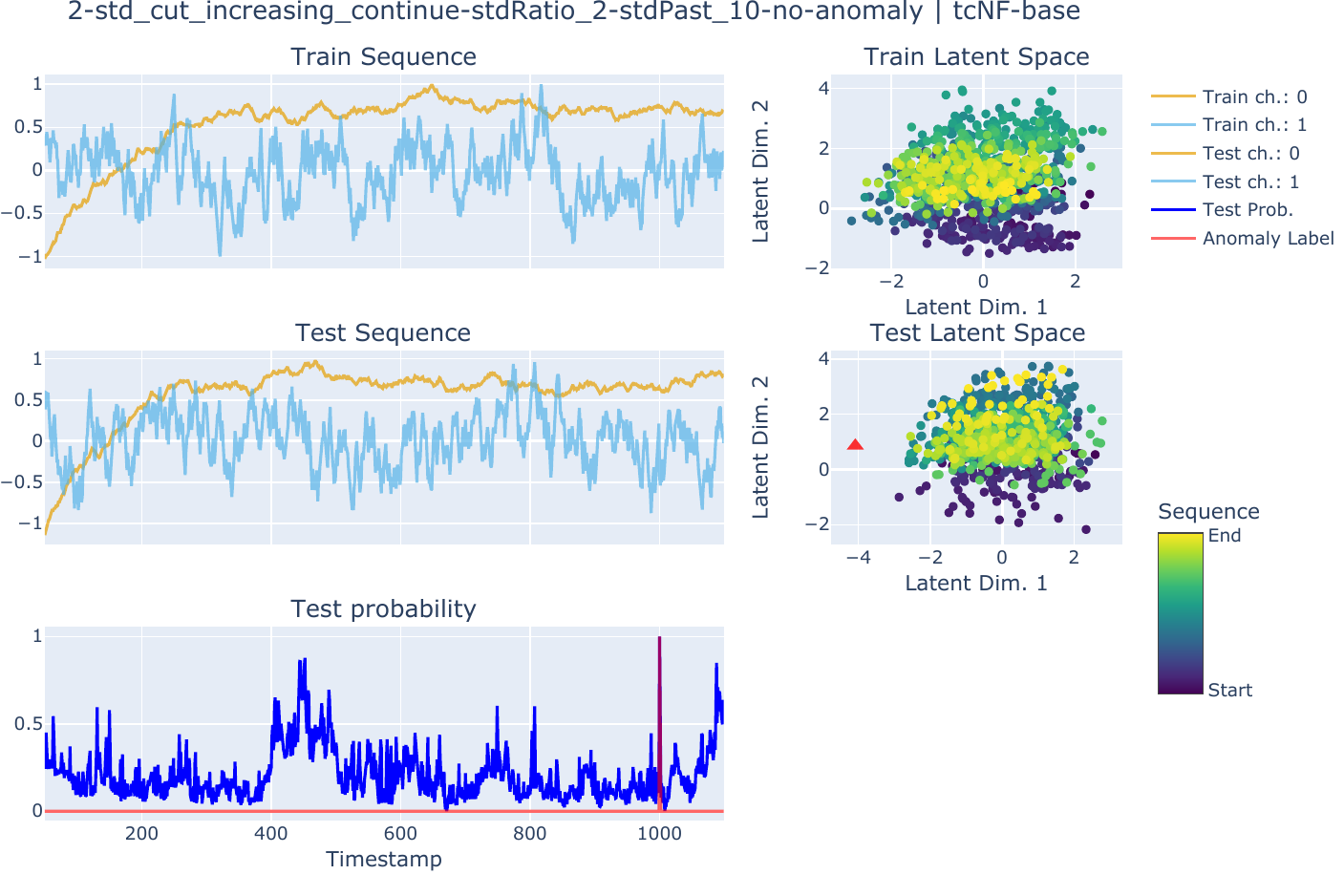}
    \vspace{-1em}
    \caption{The first row shows the training sequences and the latent representation to the right. The second row contains the test sequence with no clearly visible anomaly in the sequence to the human eye. The latent representation to the right clearly shows one out-of-distribution point as indicated by the red triangle. The anomaly in this case is a sudden jump in one timestep, as marked by the anomaly label in the third row, and is based on the standard deviation of the values in the 10 previous timesteps. Both latent space representations indicate that the model captured the sequence behavior, and only noise remains.}
    % \caption{The plots in the first row illustrate the anomaly detection based on a training sequence with no anomaly and its latent representation to the right. The second row shows the test sequence with its latent representation, including a clear out-of-distribution point where the anomaly occurs. Both latent space representations indicate that the model captured the sequence behavior, and only noise remains. The last row contains the continuous probability score with a clear detection of the anomaly and possibly an FP detection depending on the threshold.}
    \label{fig:fsb-time-latent-incr-data}
\end{figure}

Fig.\,\ref{fig:fsb-time-latent-incr-data} presents a successful anomaly detection scenario with a normally distributed latent space.
Empirical analysis revealed that not all latent spaces are like that; some retain a distinct structure\,\footref{note:see-repo}. 
tcNF methods perform especially well if anomalies are present in the training when encountering a random walk sequence.
This is likely due to the inherent nature of that sequence type, which lacks long-range structure or strong correlations, making local behavior the key detection criterion. 

\begin{figure}
    \vspace{-1em}
    \includegraphics[width=\textwidth]{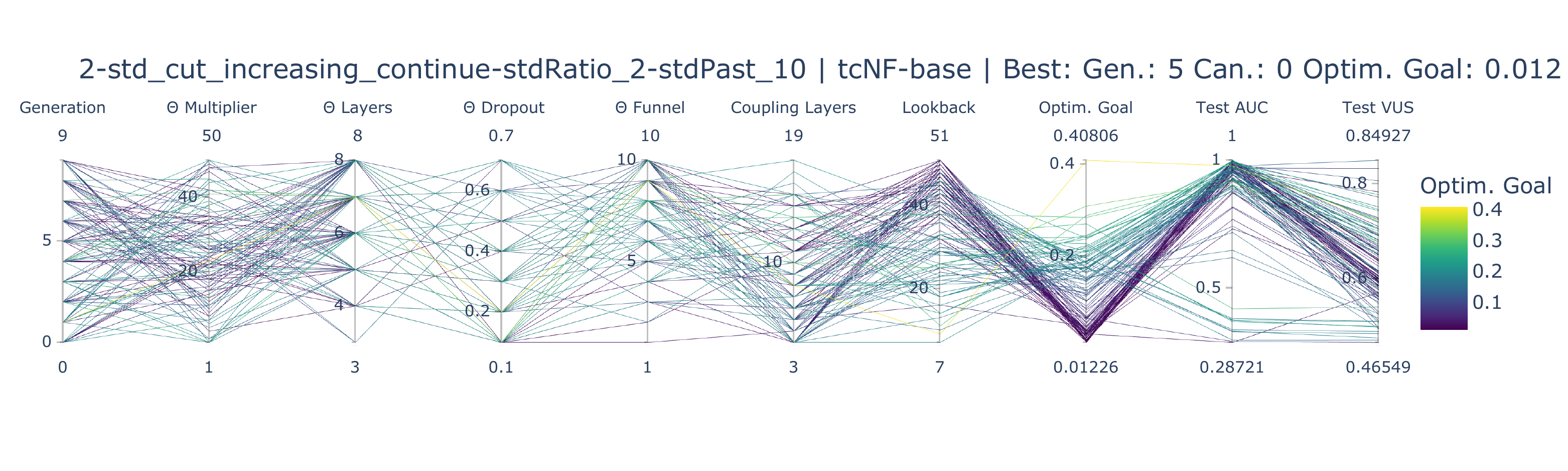}
    \vspace{-2em}
    \caption{The parameter search in this case highlights that a rather extensive access to previous datapoints is important, which correlates with an increased model capacity, such as $\Theta(\cdot)$ multiplier and $\Theta(\cdot)$ layers in each coupling layer.}
    \label{fig:fsb-parameter-space-incr-data}
\end{figure}

Fig.\,\ref{fig:fsb-parameter-space-incr-data} illustrates the full hyperparameter search space for the experiment in Fig.\,\ref{fig:fsb-time-latent-incr-data}, highlighting the relative importance of each parameter. 
The optimization process aims to minimize the training loss while maximizing the independently evaluated test accuracy.
In the given example (Fig.\,\ref{fig:fsb-parameter-space-incr-data}), performance is strongly influenced by the length of the history used and the capacity of the conditioner model, as determined by the $\Theta(\cdot)$ multiplier and number of $\Theta(\cdot)$  layers.
In contrast, regularization via $\Theta(\cdot)$  dropout and the number of coupling layers have a comparatively minor impact.

%%%%%%%%%%%%%%%%%%%%%%%%%%%%%%%%%%%%%%%%%%%%%%%%%%%%%%%%%%%%%%%%%%%%%%%%%%%%%%%%%%%%%%%%%%%%%%%%%%%%%%%%%
%%%%%%%%%%%%%%%%%%%%%%%%%%%%%%%%%%%%%%%%%%%%%%%%%%%%%%%%%%%%%%%%%%%%%%%%%%%%%%%%%%%%%%%%%%%%%%%%%%%%%%%%%
%%%%%%%%%%%%%%%%%%%%%%%%%%%%%%%%%%%%%%%%%%%%%%%%%%%%%%%%%%%%%%%%%%%%%%%%%%%%%%%%%%%%%%%%%%%%%%%%%%%%%%%%%
%%%%%%%%%%%%%%%%%%%%%%%%%%%%%%%%%%%%%%%%%%%%%%%%%%%%%%%%%%%%%%%%%%%%%%%%%%%%%%%%%%%%%%%%%%%%%%%%%%%%%%%%%
%%%%%%%%%%%%%%%%%%%%%%%%%%%%%%%%%%%%%%%%%%%%%%%%%%%%%%%%%%%%%%%%%%%%%%%%%%%%%%%%%%%%%%%%%%%%%%%%%%%%%%%%%
%%%%%%%%%%%%%%%%%%%%%%%%%%%%%%%%%%%%%%%%%%%%%%%%%%%%%%%%%%%%%%%%%%%%%%%%%%%%%%%%%%%%%%%%%%%%%%%%%%%%%%%%%
%%%%%%%%%%%%%%%%%%%%%%%%%%%%%%%%%%%%%%%%%%%%%%%%%%%%%%%%%%%%%%%%%%%%%%%%%%%%%%%%%%%%%%%%%%%%%%%%%%%%%%%%%
%%%%%%%%%%%%%%%%%%%%%%%%%%%%%%%%%%%%%%%%%%%%%%%%%%%%%%%%%%%%%%%%%%%%%%%%%%%%%%%%%%%%%%%%%%%%%%%%%%%%%%%%%

\subsection{Semi-Realistic Benchmark Suite (SRB)}
\label{subsec:results-srb}

Tab.\,\ref{tab:srb-results} presents the anomaly detection performance of tcNF models and other state-of-the-art methods on the SRB benchmark, evaluated using AUC and VUS metrics. 
Results for other methods than tcNF and RealNVP are taken from\,\cite{baumgartner_mtads_2023}.
The results demonstrate that the proposed tcNF models, particularly tcNF-base and tcNF-cnn, significantly outperform RealNVP as the methodological baseline, but is significantly poorer than IF-LOF. 
We hypothesize that IF-LOF's superiority can partly be attributed to the model being an offline technique, while the tcNF models are streaming.

Further analysis revealed several key observations regarding anomaly detection performance. 
First, aggregating point-level predictions into window-based scores or using point adjustment would improve the detection and scores. 
Second, discrepancies between detected anomalies and ground truth were primarily characterized by delayed detections or complete misses. 
Contrary to expectations, distributional shifts induced by anomalies did not consistently result in substantial probability changes by any model. 
These findings underscore the relevance and inherent challenges of the SRB suite for future research in anomaly detection.

\begin{table*}
    \centering
    \caption{Comparing anomaly detection performance on the SRB benchmark from the sequence type
    using AUC and VUS. The results show our proposed tcNF models can outperform the base model, RealNVP, but can not fully outperform all other models on this benchmark suite. See \break Tab.\,\ref{tab:fsb-results} for method categories.}
    \label{tab:srb-results}
    \scriptsize
    \begin{tabular}{lcc}
& AUC & VUS \\
Method &  &  \\
\midrule
\deepl{$\distribution$} tcNF-base & 0.73 ± 0.17 & 0.70 ± 0.20 \\
\deepl{$\distribution$} tcNF-cnn & 0.72 ± 0.19 & 0.70 ± 0.20 \\
\deepl{$\distribution$} tcNF-mlp & 0.69 ± 0.20 & 0.66 ± 0.22 \\
\deepl{$\distribution$} tcNF-stateless & 0.59 ± 0.13 & 0.54 ± 0.18 \\
\midrule
\outlier{$\trees$} IF-LOF\,\cite{cheng_outlier_2019} & \bfseries 0.90 ± 0.14 & \bfseries 0.95 ± 0.05 \\
\classic{$\distance$} KNN\,\cite{corchado_fourier_2009} & 0.68 ± 0.06 & 0.65 ± 0.13 \\
\deepl{$\distribution$} RealNVP\,\cite{dinh_density_2017} & 0.60 ± 0.11 & 0.58 ± 0.13 \\
\deepl{$\forecasting$} GDN\,\cite{deng_graph_2021} & 0.59 ± 0.19 & 0.54 ± 0.22 \\
\classic{$\distance$} HBOS\,\cite{goldstein_histogram-based_2012} & 0.58 ± 0.15 & 0.54 ± 0.18 \\
\outlier{$\trees$} iForest\,\cite{liu_isolation_2008} & 0.55 ± 0.09 & 0.51 ± 0.11 \\
\classic{$\reconstruction$} PCC\,\cite{shyu_novel_2003} & 0.51 ± 0.18 & 0.47 ± 0.16 \\
\classic{$\distance$} DAMP\,\cite{lu_damp_2023} & 0.50 ± 0.10 & 0.51 ± 0.12 \\
\outlier{$\distance$} CBLOF\,\cite{he_discovering_2003} & 0.43 ± 0.09 & 0.38 ± 0.09 \\
\classic{$\reconstruction$} PCA & 0.36 ± 0.17 & 0.32 ± 0.14 \\
\end{tabular}
   
    \vspace{-1em}
\end{table*}

%%%%%%%%%%%%%%%%%%%%%%%%%%%%%%%%%%%%%%%%%%%%%%%%%%%%%%%%%%%%%%%%%%%%%%%%%%%%%%%%%%%%%%%%%%%%%%%%%%%%%%%%%
%%%%%%%%%%%%%%%%%%%%%%%%%%%%%%%%%%%%%%%%%%%%%%%%%%%%%%%%%%%%%%%%%%%%%%%%%%%%%%%%%%%%%%%%%%%%%%%%%%%%%%%%%
%%%%%%%%%%%%%%%%%%%%%%%%%%%%%%%%%%%%%%%%%%%%%%%%%%%%%%%%%%%%%%%%%%%%%%%%%%%%%%%%%%%%%%%%%%%%%%%%%%%%%%%%%
%%%%%%%%%%%%%%%%%%%%%%%%%%%%%%%%%%%%%%%%%%%%%%%%%%%%%%%%%%%%%%%%%%%%%%%%%%%%%%%%%%%%%%%%%%%%%%%%%%%%%%%%%
%%%%%%%%%%%%%%%%%%%%%%%%%%%%%%%%%%%%%%%%%%%%%%%%%%%%%%%%%%%%%%%%%%%%%%%%%%%%%%%%%%%%%%%%%%%%%%%%%%%%%%%%%
%%%%%%%%%%%%%%%%%%%%%%%%%%%%%%%%%%%%%%%%%%%%%%%%%%%%%%%%%%%%%%%%%%%%%%%%%%%%%%%%%%%%%%%%%%%%%%%%%%%%%%%%%
%%%%%%%%%%%%%%%%%%%%%%%%%%%%%%%%%%%%%%%%%%%%%%%%%%%%%%%%%%%%%%%%%%%%%%%%%%%%%%%%%%%%%%%%%%%%%%%%%%%%%%%%%
%%%%%%%%%%%%%%%%%%%%%%%%%%%%%%%%%%%%%%%%%%%%%%%%%%%%%%%%%%%%%%%%%%%%%%%%%%%%%%%%%%%%%%%%%%%%%%%%%%%%%%%%%

\subsection{Real-World Datasets}
\label{subsec:results-real-world}

To evaluate the generalization capabilities of the proposed tcNF models, we conducted experiments on five diverse real-world datasets: SWaT, Calit2, GHL, Metro, and SMD. 
These datasets represent various domains and exhibit distinct characteristics, including rapid jumps and smoothness. 
We optimize tcNF models and RealNVP equally, resulting in models that are close to optimal for each dataset. We don't use any adjustment methods on the scoring, limiting the possible candidates with whom to compare.
Additional results are pulled from the GutenTag repository\,\footnote{\url{https://timeeval.github.io/evaluation-paper/notebooks/GutenTAG-result-analysis.html}}, and are comparable through similar scoring protocols as is the result of DeepAnT, DDANF, VTT-PAT, and MTGFlow on AUC.

Tab.\,\ref{tab:real-results} presents the anomaly detection performance on the five real-world datasets, assessed using AUC and VUS.
The results demonstrate that the tcNF models generally achieve comparable or superior performance to the RealNVP\,\cite{dinh_density_2017} baseline, particularly on the GHL dataset.
Results reported in AFNF\,\cite{guan_conditional_2023} cannot be directly compared because they use adjustment strategies for the scoring, and the code is not available to reproduce the results.
DDANF\,\cite{zhao_ddanf_2024}, on the other hand, achieves similar performance under the premise that no adjustment strategies were reported. 

Notably, the tcNF-stateful model, despite its training complexity, significantly outperforms the other normalizing flow methods on the CalIt2 dataset.
While inverting the anomaly scores was considered, our empirical analysis of the continuous probability scores confirmed that this would not resolve the performance discrepancy. 
This could mean that the optimization objectives need to be changed.
Overall, these findings suggest that tcNF models are well-suited for anomaly detection in time series, particularly on sequences with smoother temporal base patterns.

\begin{table*}
    \centering
    \caption{Comparison of anomaly detection performance on five real-world datasets using AUC (a) and VUS (b). The results demonstrate various performance improvements over the base method. SWaT, CalIt, and Metro consist of rapidly jumping values, which are difficult for our method. GHL and SMD consist of a mix of jumps and smoother channels, which work better for our method. See Tab.\,\ref{tab:fsb-results} for method categories.}
    \label{tab:real-results}
    \scriptsize
        \begin{tabular}{lccccc}
\multicolumn{6}{c}{(a) \textbf{AUC}} \\[0.2cm]
& SWaT & CalIt & GHL & Metro & SMD \\
Method &  &  &  &  &  \\
\midrule
\deepl{$\distribution$} tcNF-base & \bfseries 0.88 & 0.25 & 0.88 ± 0.09 & 0.48 & 0.78 ± 0.11 \\
\deepl{$\distribution$} tcNF-mlp & \bfseries 0.88 & 0.30 & 0.64 ± 0.34 & 0.67 & 0.80 ± 0.14 \\
\deepl{$\distribution$} tcNF-cnn & 0.84 & 0.29 & \bfseries 0.89 ± 0.04 & 0.49 & 0.80 ± 0.13 \\
\deepl{$\distribution$} tcNF-stateless & \bfseries 0.88 & 0.23 & 0.58 ± 0.40 & 0.53 & 0.78 ± 0.14 \\
\deepl{$\distribution$} tcNF-stateful & - & 0.76 & - & - & - \\
\midrule
\deepl{$\distribution$} RealNVP\,\cite{dinh_density_2017} & \bfseries 0.88 & 0.22 & 0.69 ± 0.22 & \bfseries 0.69 & 0.80 ± 0.12 \\
\classic{$\distance$} k-Means\,\cite{yairi_fault_2001} & - & \bfseries 0.92 & - & - & \bfseries 0.97 ± 0.02 \\
\classic{$\distance$} KNN\,\cite{corchado_fourier_2009} & - & 0.88 & - & - & 0.85 ± 0.06 \\
\outlier{$\trees$} iForest\,\cite{liu_isolation_2008} & - & 0.88 & - & - & 0.85 ± 0.07 \\
\classic{$\reconstruction$} PCC\,\cite{shyu_novel_2003} & - & 0.75 & - & - & 0.82 ± 0.08 \\
\deepl{$\forecasting$} DeepAnT\,\cite{munir_deepant_2019} & - & - & - & - & 0.78 ± 0.16 \\
%\deepl{$\distribution$} AFNF & \bfseries 0.90 & - & - & - & 0.96 \\
\deepl{$\reconstruction$} DDANF\,\cite{zhao_ddanf_2024} & \bfseries 0.88 & - & - & - & - \\
\deepl{$\reconstruction$} VTT-PAT\,\cite{kang_transformer-based_2024} & 0.84 & - & - & - & 0.57 \\
% \deepl{$\reconstruction$} USAD & 0.81 & - & - & - & 0.57 \\
\deepl{$\distribution$} MTGFlow\,\cite{zhou_label-free_2024} & 0.85 & - & - & - & 0.91 \\
\end{tabular}

        \vspace{-8pt}
        \begin{tabular}{lccccc}
\multicolumn{6}{c}{(b) \textbf{VUS}} \\[0.2cm]
& SWaT & CalIt & GHL & Metro & SMD \\
Method &  &  &  &  &  \\
\midrule
\deepl{$\distribution$} tcNF-base & \bfseries 0.88 & 0.40 & 0.89 ± 0.09 & 0.78 & 0.80 ± 0.09 \\
\deepl{$\distribution$} tcNF-mlp & \bfseries 0.88 & 0.44 & 0.66 ± 0.32 & 0.80 & 0.80 ± 0.12 \\
\deepl{$\distribution$} tcNF-cnn & 0.83 & 0.44 & \bfseries 0.90 ± 0.04 & 0.76 & 0.80 ± 0.11 \\
\deepl{$\distribution$} tcNF-stateless & 0.86 & 0.35 & 0.59 ± 0.38 & 0.79 & 0.80 ± 0.11 \\
\deepl{$\distribution$} tcNF-stateful & - & \bfseries 0.77 & - & - & - \\
\deepl{$\distribution$} RealNVP\cite{dinh_density_2017} & \bfseries 0.88 & 0.35 & 0.70 ± 0.22 & \bfseries 0.85 & \bfseries 0.81 ± 0.10 \\
\end{tabular}

    \vspace{-2em}
\end{table*}

\begin{figure}
    \includegraphics[width=\textwidth]{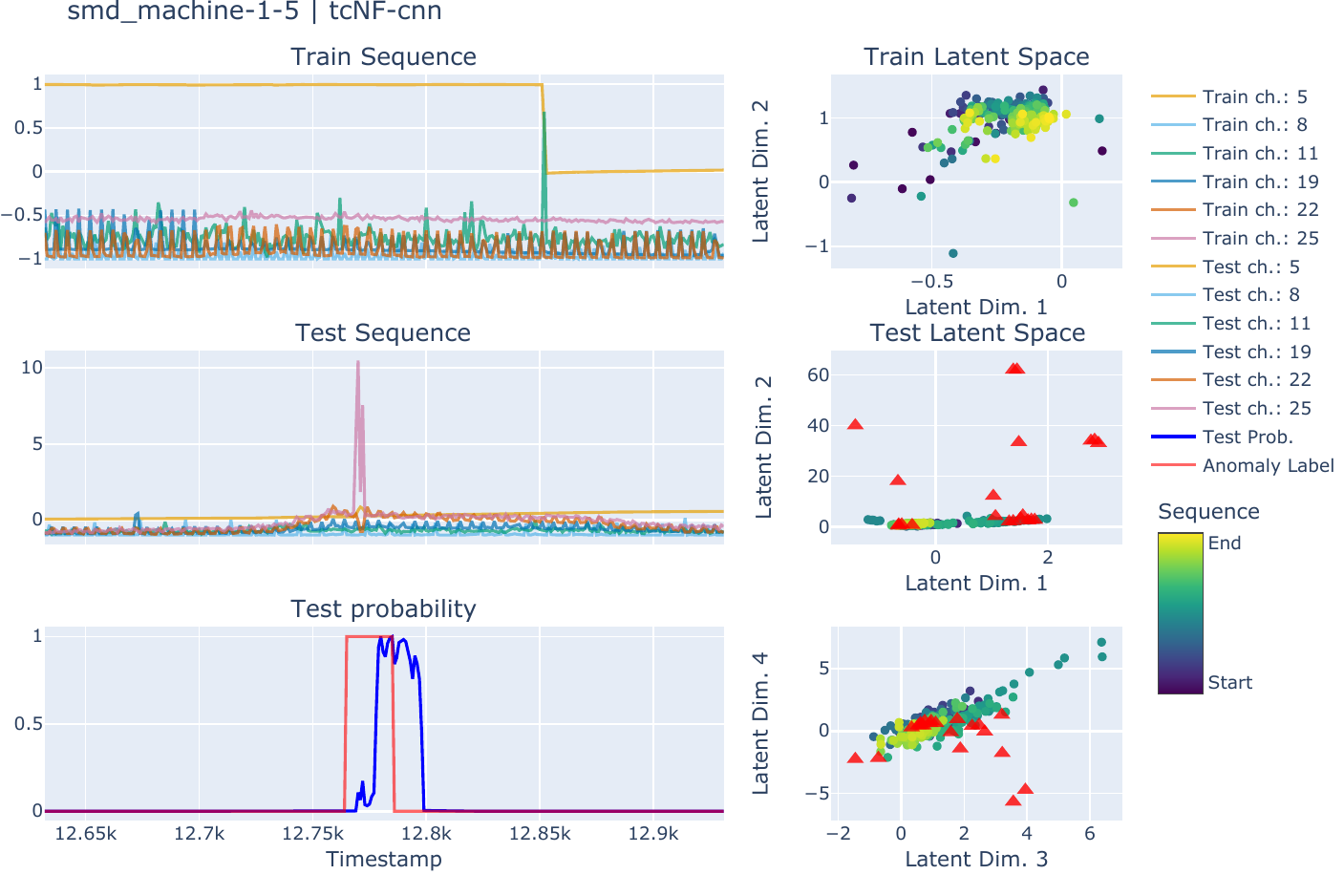}
    \vspace{-1em}
    \caption{Clip-out of the machine-1-5 sequence in the SMD dataset showing the training sequence with some normal and unusual behaviour. The test sequence contains a strong anomaly in two out of the 38 channels, where six distinct channels are shown. This case shows the delay of the tcNF-cnn method, which relies more strongly on historical behavior via the conditioning.}
    \label{fig:real-tcNF-cnn-smd-1-5-latentspace}
\end{figure}

\begin{figure}
    \vspace{-1em}
    \includegraphics[width=\textwidth]{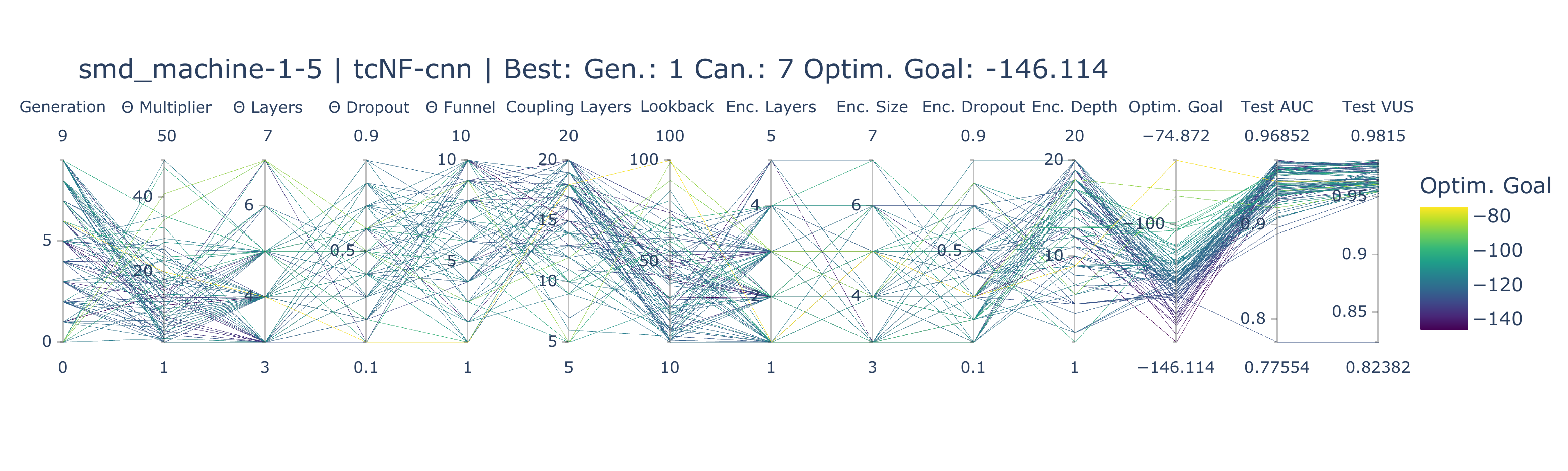}
    \vspace{-1em}
    \caption{The most significant parameters in this example are the $\Theta(\cdot)$  multiplier, $\Theta(\cdot)$  layers, $\Theta(\cdot)$  funnel, number of coupling layers and look-back size. The  parameters indicate a successful regularisation by the optimization search and the other two indicate that a certain amount is needed to be able to learn a good representation.}
    \label{fig:real-tcNF-cnn-smd-1-5-paramsearch}
\end{figure}

Fig.\,\ref{fig:real-tcNF-cnn-smd-1-5-latentspace} and \ref{fig:real-tcNF-cnn-smd-1-5-paramsearch} present results demonstrating success in anomaly detection using the tcNF-cnn model, revealing specific model behavior. 
Time series plots (Fig.\,\ref{fig:real-tcNF-cnn-smd-1-5-latentspace}) display six representative channels (out of 38), exhibiting both typical and anomalous behavior.
The latent space from the training data shows that the remaining information not captured by the model is scattered around as expected. 
The test latent space highlights clearly that unknown behavior is less likely. 

The tcNF-cnn architecture uses a CNN encoder that connects consecutive data points.
This can lead to detection delay, resulting in false negatives (FNs) at the anomaly beginning and false positives (FPs) immediately following the anomalous period.
This behavior is also reflected in the latent space (Fig. \ref{fig:real-tcNF-cnn-smd-1-5-latentspace}), where anomalous data points are displaced from the distribution's center, with varying magnitudes across different dimensions and within.

The parameter search plot (Fig. \ref{fig:real-tcNF-cnn-smd-1-5-paramsearch}) identifies four key parameters: the $\Theta(\cdot)$  multiplier, the number of $\Theta(\cdot)$  layers, the $\Theta(\cdot)$  funneling factor, and the number of coupling layers.
Results indicate that a conditioner function with few layers, a substantial funneling effect, and a sufficient number of coupling layers, yields a suitable model configuration for this specific scenario.


\begin{figure}
    \includegraphics[width=\textwidth]{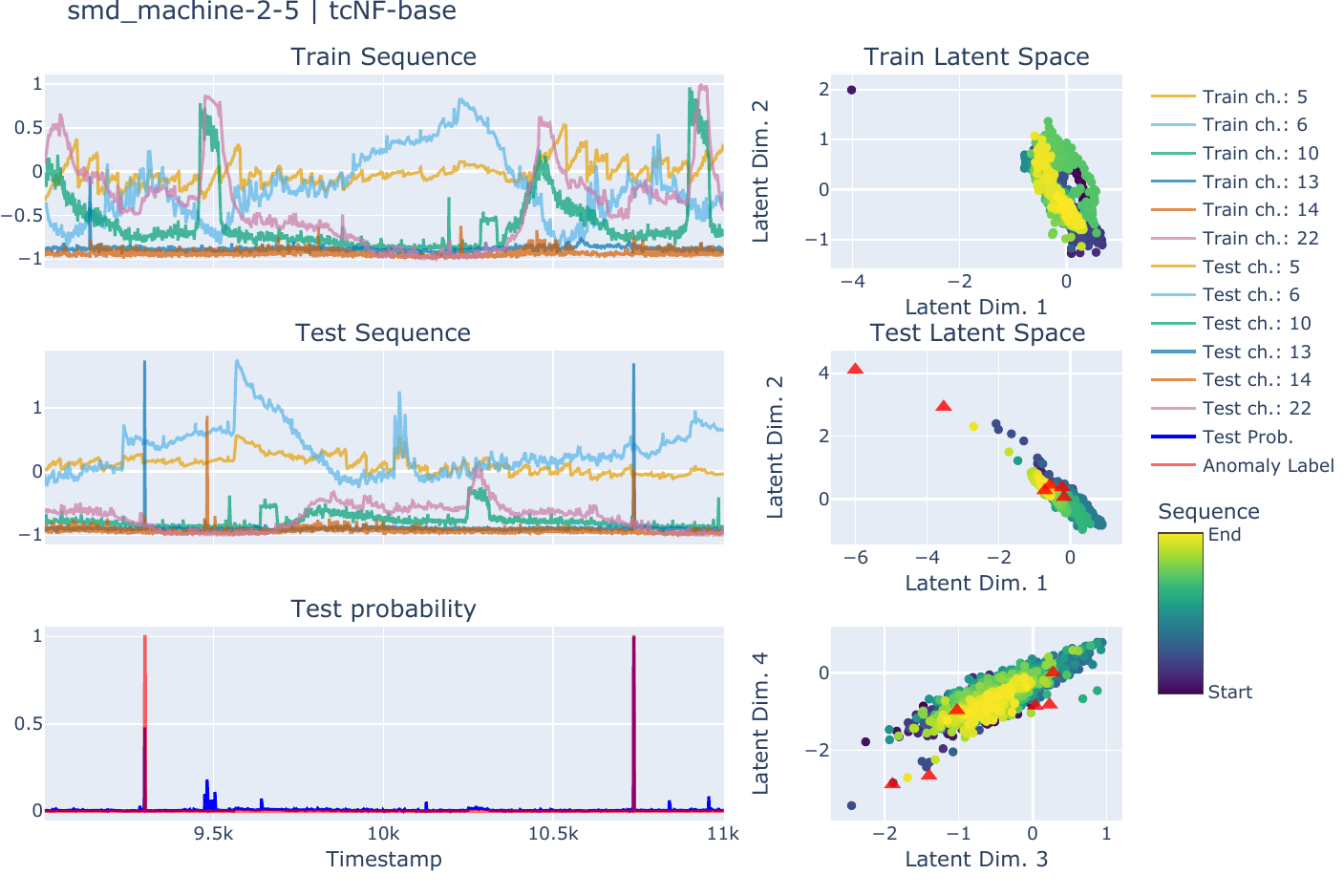}
    \vspace{-1em}
    \caption{Clip-out of the machine-2-5 sequence in the SMD dataset showing a normal training set with one exception in the beginning which can also be seen in the latent space as outlier. The test sequence contains two labeled anomalies. A zoom in would reveal an early detection meaning at the beginning of the anomaly, which indicates that, in this case, the near past is important for the detection.}
    \label{fig:real-tcNF-baes-smd-2-5-latentspace}
\end{figure}

\begin{figure}
    \vspace{-1em}
    \includegraphics[width=\textwidth]{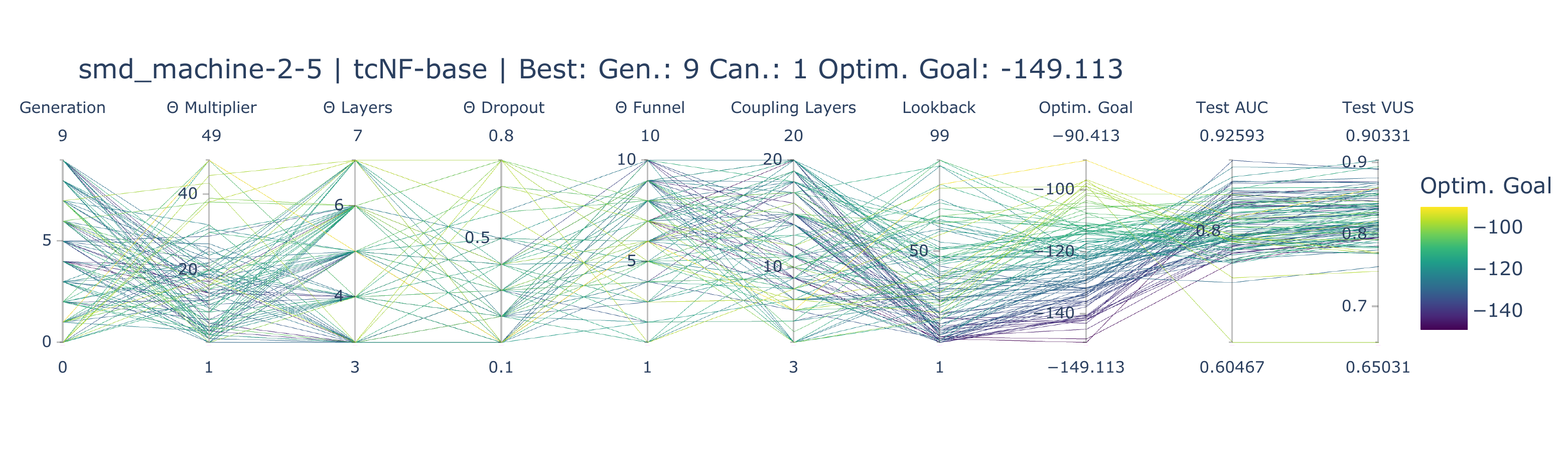}
    \vspace{-2em}
    \caption{The parameter optimization search reveals exactly that the previous observations is important for this sequence, including some regularisation on the other parameters.}
    \label{fig:real-tcNF-base-smd-2-5-paramsearch}
\end{figure}

Fig.\,\ref{fig:real-tcNF-baes-smd-2-5-latentspace} and \ref{fig:real-tcNF-base-smd-2-5-paramsearch} demonstrate varying anomaly detection performance using the tcNF-base model, highlighting specific model characteristics.
Fig.\,\ref{fig:real-tcNF-baes-smd-2-5-latentspace} displays time series plots for six representative channels (selected from 38) exhibiting periods of nominal behavior.
The corresponding latent space representation suggests a near-normal distribution, indicating that the model does not fully capture all observed behaviors.
Notably, a single outlier is observed in the top-left corner of the latent space, corresponding to a measurement spike at the beginning of the train sequence in channel 13.

The tcNF-base architecture prioritizes early anomaly detection, relying heavily on the current input and recent observations.
This characteristic leads to a rapid decrease in anomaly detection probability and, consequently, some missed detections (FPs), as evidenced in the latent space of the test sequence. 
Furthermore, channel 14 exhibits an increase in anomaly probability near timestamp 9500, suggesting a period of subtle anomalous behavior.

The parameter search results (Fig.\,\ref{fig:real-tcNF-base-smd-2-5-paramsearch}) are qualitatively similar to those presented in Fig.\,\ref{fig:real-tcNF-cnn-smd-1-5-paramsearch}; however, in this case, utilizing a shorter history window yields improved performance.
While the current solution was identified in the final generation of the search, it is plausible that extending the search with additional candidates and generations could have yielded further performance improvements.


%%%%%%%%%%%%%%%%%%%%%%%%%%%%%%%%%%%%%%%%%%%%%%%%%%%%%%%%%%%%%%%%%%%%%%%%%%%%%%%%%%%%%%%%%%%%%%%%%%%%%%%%%
%%%%%%%%%%%%%%%%%%%%%%%%%%%%%%%%%%%%%%%%%%%%%%%%%%%%%%%%%%%%%%%%%%%%%%%%%%%%%%%%%%%%%%%%%%%%%%%%%%%%%%%%%
%%%%%%%%%%%%%%%%%%%%%%%%%%%%%%%%%%%%%%%%%%%%%%%%%%%%%%%%%%%%%%%%%%%%%%%%%%%%%%%%%%%%%%%%%%%%%%%%%%%%%%%%%
%%%%%%%%%%%%%%%%%%%%%%%%%%%%%%%%%%%%%%%%%%%%%%%%%%%%%%%%%%%%%%%%%%%%%%%%%%%%%%%%%%%%%%%%%%%%%%%%%%%%%%%%%
%%%%%%%%%%%%%%%%%%%%%%%%%%%%%%%%%%%%%%%%%%%%%%%%%%%%%%%%%%%%%%%%%%%%%%%%%%%%%%%%%%%%%%%%%%%%%%%%%%%%%%%%%
%%%%%%%%%%%%%%%%%%%%%%%%%%%%%%%%%%%%%%%%%%%%%%%%%%%%%%%%%%%%%%%%%%%%%%%%%%%%%%%%%%%%%%%%%%%%%%%%%%%%%%%%%
%%%%%%%%%%%%%%%%%%%%%%%%%%%%%%%%%%%%%%%%%%%%%%%%%%%%%%%%%%%%%%%%%%%%%%%%%%%%%%%%%%%%%%%%%%%%%%%%%%%%%%%%%
%%%%%%%%%%%%%%%%%%%%%%%%%%%%%%%%%%%%%%%%%%%%%%%%%%%%%%%%%%%%%%%%%%%%%%%%%%%%%%%%%%%%%%%%%%%%%%%%%%%%%%%%%

\subsection{Discussion}
\label{subsec:discussion}

Extensive experiments across multiple benchmark datasets and anomaly detection scenarios demonstrate the robust anomaly detection capabilities of our unsupervised tcNF framework. 
It consistently outperforms the RealNVP baseline method and is highly competitive against the state-of-the-art methods.
These experiments also highlight the flexibility of tcNF, enabling its adaptation to specific application needs and the benefit of the automated framework to find a suitable solution for a given dataset.
The characteristics of our framework make it ideal as a simple-to-test solution that can reach initial results with minimal effort. 

Our experiments encompass a broad spectrum of time series data, ranging from low- to high-dimensional scenarios.
The results indicate that shorter sequences that contain high-quality representations of normal behavior are generally more advantageous than longer sequences that may potentially be contaminated by unidentified anomalies.
In line with the definition of anomaly detection as described in Sec.\,\ref{sec:introduction}, extensive appearances of anomalies in the training data are inherently challenging to detect in subsequent test scenarios. 
Our framework effectively addresses this challenge in many cases. 
In fact, there are specific scenarios where incorporating anomalies as noise in the training data can be beneficial, yielding better detection performance.
This is especially the case when the model faces sequences with not enough variation or a lack of structure in the data, such as in the cylinder-bell-funnel (CBF) or random walk (RW) sequences, leading to overfitting or failing to learn a useful representation.

\section{Conclusions}
\label{sec:conclusion}

This work introduced a novel unsupervised probabilistic framework for anomaly detection in multivariate time series. 
By directly conditioning a normalizing flow on relevant information, we gained a flexible and efficient approach. 
The end-to-end training process enabled seamless optimization, including hyperparameter tuning.
Our experiments demonstrated the framework's robustness and competitive performance compared to existing methods, particularly in terms of computational efficiency. 
Nevertheless, the presence of numerous anomalies in the training data can significantly impact performance. 
To address this challenge, we suggested an analysis of the posterior distribution of the training data to identify and mitigate the influence of such anomalies, especially when they occur infrequently.
To conclude, our proposed framework provides a promising foundation for anomaly detection in multivariate time series, especially the inference process for real-time detection is significantly more efficient, compared to other generative models. 

More research is needed to address the limitations we identified through this work. 
The following are directions for future research that we propose to enhance the capabilities and broaden the applicability our approach in various domains:

\begin{itemize}
    \item \textbf{Improving the conditioning mechanism:} Exploring and extending more sophisticated conditioning strategies, such as incorporating transformers, or leveraging local window normalization as proposed in \cite{rasul_lag-llama_2024}, could enhance the model's adaptability and detection accuracy.
    \item \textbf{Impact of noise/anomalies in training:} Exploring the ratio between expected and unexpected data in contracts to the detection is an important aspect to follow for this and other unsupervised methods.
    \item \textbf{Understanding the detection cause:} Investigating methods to provide insights into the specific factors contributing to an anomaly detection is crucial for interpretability and trust in real-world applications.
    \item \textbf{Real-world anomaly detection dataset:} This work and the referenced, highlight the need for a fair real-world dataset for anomaly detection, including guidelines on calculating the detection scores.
\end{itemize}

%%%%%%%%%%%%%%%

\backmatter

\bmhead{Acknowledgements}

This work has been carried out at the Centre for Research-based Innovation, SFI NorwAI, funded by the Research Council of Norway under grant no. 309834.

%\begin{appendices}

% \input{content/appendix}

%\end{appendices}

\bibliography{mybibfile}

\end{document}